\documentclass[conference]{IEEEtran}
\IEEEoverridecommandlockouts
\usepackage{cite}
\usepackage{amsmath,amssymb,amsfonts}
\usepackage{algorithmic}
\usepackage{graphicx}
\usepackage{textcomp}
\usepackage{xcolor}

\usepackage{hyperref}
\usepackage{xspace}
\usepackage{multirow}
\usepackage{subcaption}
\usepackage{makecell}
\usepackage{balance}
\usepackage{siunitx}
\usepackage{booktabs}
\usepackage{mathtools}

\newcommand{\set}[1]{\mathcal{#1}}
\providecommand{\sE}{\ensuremath{\set{E}}}
\providecommand{\sR}{\ensuremath{\set{R}}}
\providecommand{\sV}{\ensuremath{\set{V}}}
\providecommand{\sN}{\ensuremath{\set{N}}}
\providecommand{\sL}{\ensuremath{\set{L}}}
\providecommand{\sS}{\ensuremath{\set{S}}}

\renewcommand{\vec}[1]{{\bf{#1}}}
\providecommand{\va}{\ensuremath{\vec{a}}}
\providecommand{\vh}{\ensuremath{\vec{h}}}
\providecommand{\vx}{\ensuremath{\vec{x}}}
\providecommand{\vy}{\ensuremath{\vec{y}}}

\providecommand{\vba}{\ensuremath{\overline{\vec{a}}}}
\providecommand{\vta}{\ensuremath{\widetilde{\vec{a}}}}

\newcommand{\mat}[1]{\boldsymbol{#1}}
\providecommand{\mW}{\ensuremath{\mat{W}}}
\providecommand{\mX}{\ensuremath{\mat{X}}}

\providecommand{\mP}{\ensuremath{\mat{P}}}
\providecommand{\mbP}{\ensuremath{\overline{\mat{P}}}}
\providecommand{\mbW}{\ensuremath{\overline{\mat{W}}}}
\providecommand{\mtP}{\ensuremath{\widetilde{\mat{P}}}}
\providecommand{\mtW}{\ensuremath{\widetilde{\mat{W}}}}

\providecommand{\gammail}{\ensuremath{\text{\Large{$\gamma$}}_{il}}}

\newcommand{\amazon}[0]{\textsl{Amazon}\xspace}
\newcommand{\yelp}[0]{\textsl{YelpChi}\xspace}

\newcommand{\ours}[0]{\textup{DRAG}\xspace}
\newcommand{\gsage}[0]{\textup{GraphSAGE}\xspace}
\newcommand{\gat}[0]{\textup{GAT}\xspace}
\newcommand{\gattwo}[0]{\textup{GATv2}\xspace}
\newcommand{\bwgnn}[0]{\textup{BWGNN}\xspace}
\newcommand{\bwgnnhomo}[0]{\textup{BWGNN-Homo}\xspace}
\newcommand{\bwgnnhete}[0]{\textup{BWGNN-Hetero}\xspace}
\newcommand{\fraudre}[0]{\textup{FRAUDRE}\xspace}
\newcommand{\pcgnn}[0]{\textup{PC-GNN}\xspace}
\newcommand{\caregnn}[0]{\textup{CARE-GNN}\xspace}
\newcommand{\mlp}[0]{\textup{MLP}\xspace}

\newcommand{\fone}[0]{F1-macro\xspace}
\newcommand{\auc}[0]{AUC\xspace}

\newcommand{\nummodels}[0]{10}

\def\BibTeX{{\rm B\kern-.05em{\sc i\kern-.025em b}\kern-.08em
    T\kern-.1667em\lower.7ex\hbox{E}\kern-.125emX}}
\begin{document}

\title{Dynamic Relation-Attentive Graph Neural Networks for Fraud Detection
 \thanks{This work was supported by the National Research Foundation of Korea grant funded by MSIT (2022R1A2C4001594 and 2018R1A5A1059921).}
}

\author{\IEEEauthorblockN{Heehyeon Kim}
\IEEEauthorblockA{\textit{School of Computing} \\
\textit{KAIST}\\
Daejeon, Republic of Korea \\
heehyeon@kaist.ac.kr}
\and
\IEEEauthorblockN{Jinhyeok Choi}
\IEEEauthorblockA{\textit{School of Computing} \\
\textit{KAIST}\\
Daejeon, Republic of Korea \\
cjh0507@kaist.ac.kr}
\and
\IEEEauthorblockN{Joyce Jiyoung Whang$^{\dagger}$ \thanks{$^{\dagger}$Corresponding author.}}
\IEEEauthorblockA{\textit{School of Computing} \\
\textit{KAIST}\\
Daejeon, Republic of Korea \\
jjwhang@kaist.ac.kr}
}
\maketitle

\begin{abstract}
Fraud detection aims to discover fraudsters deceiving other users by, for example, leaving fake reviews or making abnormal transactions. Graph-based fraud detection methods consider this task as a classification problem with two classes: frauds or normal. We address this problem using Graph Neural Networks (GNNs) by proposing a dynamic relation-attentive aggregation mechanism. Based on the observation that many real-world graphs include different types of relations, we propose to learn a node representation per relation and aggregate the node representations using a learnable attention function that assigns a different attention coefficient to each relation. Furthermore, we combine the node representations from different layers to consider both the local and global structures of a target node, which is beneficial to improving the performance of fraud detection on graphs with heterophily. By employing dynamic graph attention in all the aggregation processes, our method adaptively computes the attention coefficients for each node. Experimental results show that our method, DRAG, outperforms state-of-the-art fraud detection methods on real-world benchmark datasets.
\end{abstract}

\section{Introduction}
Graph-based fraud detection methods, also called graph anomaly detection methods, represent objects that should be determined to be fraud or benign as nodes and make edges between them~\cite{survey,bond}. For example, in \yelp benchmark dataset~\cite{yelp}, nodes are reviews and edges are created based on three different factors: (1) whether the reviews were written by the same user, (2) whether the reviews were written in the same month, and (3) whether the reviews had the same star rating. Each of these three factors can be considered as a different relation since their semantics are distinct~\cite{fake}.

Several recently proposed fraud detection methods distinguish different relations in computing node representations~\cite{ngs,dagnn,h2fd,gconsis,gdn,aognn}. For example, \caregnn~\cite{caregnn} uses a relation-aware neighbor aggregator and \bwgnn~\cite{bwgnn} performs a propagation process for each relation and apply a maximum pooling. Also, \fraudre~\cite{fraudre} learns the fraud-aware graph convolution model under each relation. In general, these relation-aware approaches have shown superior performance over the methods that ignore relations and consider all edges equally~\cite{gccad,amnet}.

\begin{figure*}[t]
\includegraphics[width=2\columnwidth]{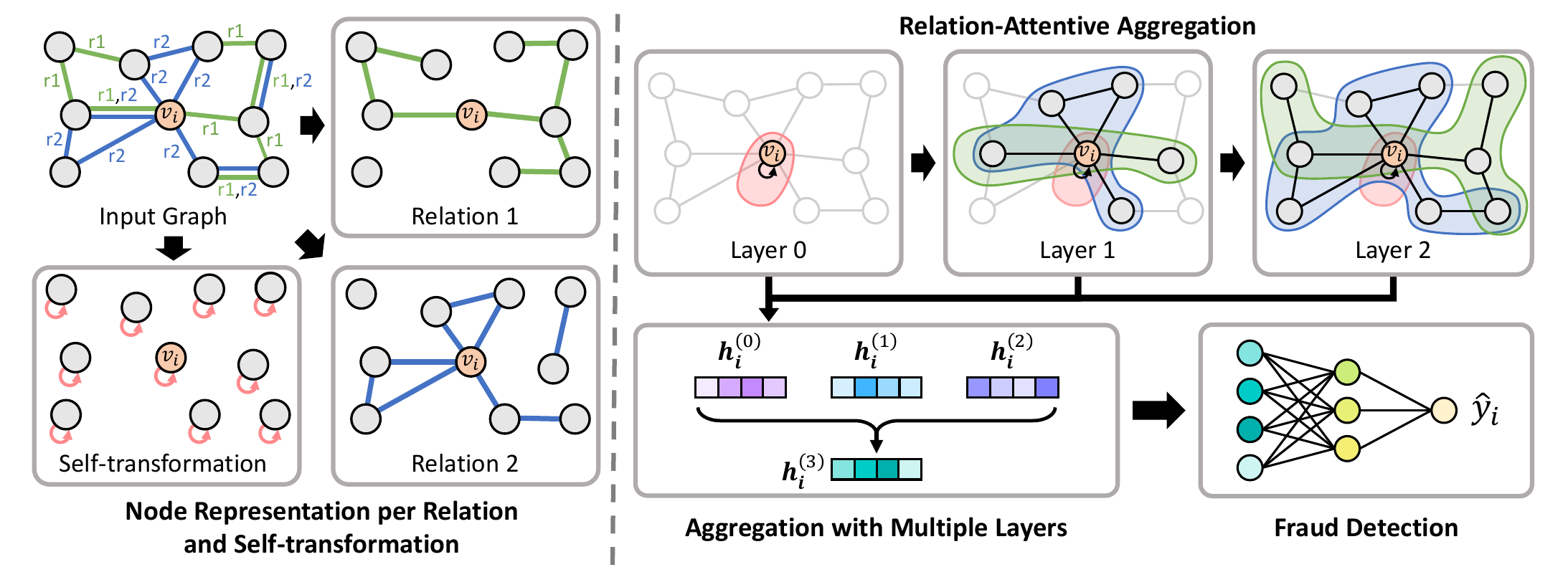}
\caption{Overview of \ours. A node representation is computed by each relation and by a self-transformation. Using learnable attention weights, the node representations are aggregated over the relations. The final node representation is learned by aggregating the representations from different layers and used to predict whether the node is a fraud.}
\label{fig:drag}
\vspace{-0.3cm}
\end{figure*}

In this paper, we propose \ours which is a \underline{D}ynamic \underline{R}elation-\underline{A}ttentive \underline{G}raph neural network (Figure~\ref{fig:drag}), which decomposes the original graph by relations to learn a node representation per relation along with a self-transformation, resulting in multiple representations for each node. We consider the self-loop used in self-transformation as another relation. At each layer, \ours aggregates the multiple representations for each node with different learnable attention weights for the relations. The final representation is computed by aggregating representations from different layers, where not only the last layer's representation but also intermediate layers' representations are taken into account. In all these processes, we employ a dynamic graph attention mechanism~\cite{gat2} to let \ours have various distributions of attention coefficients, which can differ for each node. Experimental results on real-world datasets show that \ours outperforms eight different baseline methods. Our implementations and datasets are available at \url{https://github.com/bdi-lab/DRAG}.

\paragraph*{\noindent \bf{Related Work}}
In the original GAT~\cite{gat}, the static attention problem has been issued, a phenomenon in which all nodes have a fixed ranking of attention coefficients. To resolve this issue, GATv2~\cite{gat2} has been proposed by introducing a dynamic attention, which swaps the order of operations of applying a linear projection layer and the non-linear function. On the other hand, WIRGAT~\cite{rgat} has been proposed to compute relation-wise representations for each node using a static attention and simply sum over the node representations. Different from \ours, none of these methods considers dynamic relation-attentive and layer-attentive aggregations.

Heterophily in graphs has been considered as a challenging issue in graph-based fraud detection~\cite{hetero,acm,ccns} since nodes are connected to other nodes belonging to a different class. \ours alleviates this issue by learning attention coefficients, which weigh the importance of each neighbor regarding computing the representation of the target node. More importantly, these attention coefficients can vary depending on each target node since we utilize the dynamic attention mechanism.

\section{Problem Definition}
\label{sec:def}
Let us consider an undirected multi-relation graph $G=(\sV,\sR,\sE,\mX)$ where $\sV$ is a set of nodes, $\sR$ is a set of relations, $|\sV|=n$, $|\sR|=m$, $\sE=\{(v_i,r_k,v_j)|v_i\in\sV,r_k\in\sR,v_j\in\sV\}$, $\mX\in\mathbb{R}^{n\times d}$ is a set of features for the nodes, and $d$ is the dimension of each feature vector. There can be multiple edges between a pair of nodes if they have different relations. For example, given $v_i$ and $v_j$, $\sE$ can include both $(v_i,r_{k},v_j)$ and $(v_i,r_{k'},v_j)$. Each node is labeled as a normal node or a fraudulent node. Let $\vy\in\{0,1\}^{n}$ denote a vector for the labels such that $y_i=1$ if the $i$-th node is a fraud and $y_i=0$ otherwise. 

\section{\ours: Dynamic Relation-Attentive Graph Neural Network}
We describe \ours which computes node representations using relation-wise and layer-wise dynamic attention mechanisms.


\subsection{Node Representation per Relation and Self-Transformation}
\label{sec:perrel}
We propose learning multiple node representations for each node by computing a representation per relation and a self-transformation. To appropriately compute an attention coefficient~\cite{gat}, we add a self-loop to each node for all relations. The neighborhood set of $v_i$ for $r_k$ is defined by $\sN_{ik}\coloneqq\{v_j|(v_j,r_k,v_i)\in\sE\}$ where $i=1,\cdots,n$ and $k=1,\cdots,m$. The node representation of $v_i$ for $r_k$ at the $l$-th layer is denoted by $\vh_{i,k}^{(l)}\in\mathbb{R}^{d^{\prime}}$ where $d^{\prime}$ is the dimension of the node representation. Let $\vh_{i}^{(l)}\in\mathbb{R}^{d^{\prime}}$ be the node representation of $v_i$ at the $l$-th layer. We compute $\vh_{i}^{(0)}=\mlp(\vx_i)$ for $v_i\in\sV$ where \mlp is a multi-layer perceptron. Using the dynamic multi-head attention~\cite{gat2} with $N_{\alpha}$ heads, we compute the node representation for $r_k$ at each head as follows:
\begin{equation}\label{eq:perr}
\vh_{i,k}^{(l)} = \sigma\left(\sum_{v_{j}\in\sN_{ik}} \alpha_{ijk}^{(l)} \mP_k^{(l)} \vh_{j}^{(l)} \right)
\end{equation}
where $\mP_k^{(l)}\in\mathbb{R}^{(d^{\prime}/N_{\alpha})\times d^{\prime}}$ and $\alpha_{ijk}^{(l)}$ is computed by
\begin{equation}
\alpha_{ijk}^{(l)} = \dfrac{\text{exp}\left(\va_k^{(l)} \sigma\left(\mW^{(l)}_k [\vh_{i}^{(l)}||\text{ }\vh_{j}^{(l)}]\right)\right)}{\sum_{v_{j'}\in\sN_{ik}}\text{exp}\left(\va_k^{(l)} \sigma\left(\mW^{(l)}_k [\vh_{i}^{(l)}||\text{ }\vh_{j'}^{(l)}]\right)\right)}
\end{equation}
where $||$ denotes a vertical concatenation, $\mW^{(l)}_k\in\mathbb{R}^{d^{\prime}\times 2d^{\prime}}$, $\va_k^{(l)}\in\mathbb{R}^{d^{\prime}}$, $\sigma(\cdot)$ is a non-linear function, $l=0,1,\cdots,L$, and $L$ is the total number of layers. To aggregate the outputs from $N_{\alpha}$ heads, we concatenate the resulting representations from different heads~\cite{att}; as a result, we have $\vh_{i,k}^{(l)}\in\mathbb{R}^{d^{\prime}}$. We apply the same concatenation strategy to aggregate results from multi-head outputs in all following attention mechanisms.

By computing a node representation per relation using (\ref{eq:perr}), we have $m$ representations for each node. We also compute the ($m+1$)-th representation for a node by considering a self-transformation: $\vh_{i,m+1}^{(l)} = \mlp \left(\vh_{i}^{(l)}\right)$ where $\vh_{i,m+1}^{(l)}\in\mathbb{R}^{d^{\prime}}$.


\begin{table*}[t]
\footnotesize
\renewcommand{\arraystretch}{1.2}
\caption{Fraud detection performance on \yelp and \amazon, using different percentages of labels.}
\setlength{\tabcolsep}{0.84em}
\label{tb:mainexp}
\centering
\begin{tabular}{cc|cc|cc|cc}
\Xhline{2\arrayrulewidth}
\multirow{2}{*}{} & & \multicolumn{2}{c|}{1\%}  & \multicolumn{2}{c|}{10\%}  & \multicolumn{2}{c}{40\%} \\
&  & \multicolumn{1}{c}{\fone} & \multicolumn{1}{c|}{\auc} & \multicolumn{1}{c}{\fone} & \multicolumn{1}{c|}{\auc} & \multicolumn{1}{c}{\fone} & \multicolumn{1}{c}{\auc} \\
\Xhline{\arrayrulewidth}
 \multirow{\nummodels}{*}{\yelp} & \mlp & 0.6150$\pm$0.0072 & 0.7253$\pm$0.0098 & 0.6720$\pm$0.0069 & 0.8010$\pm$0.0053 & 0.7140$\pm$0.0048 & 0.8489$\pm$0.0064 \\
&\gsage & 0.6269$\pm$0.0107 & 0.7363$\pm$0.0145 & 0.6971$\pm$0.0074 & 0.8293$\pm$0.0058 & 0.7456$\pm$0.0078 & 0.8800$\pm$0.0061 \\
&\gat & 0.6183$\pm$0.0133 & 0.7188$\pm$0.0176 & 0.6763$\pm$0.0089 & 0.7970$\pm$0.0118 & 0.7190$\pm$0.0085 & 0.8492$\pm$0.0087 \\
&\gattwo & 0.6283$\pm$0.0109 & 0.7366$\pm$0.0129 & 0.6938$\pm$0.0077 & 0.8204$\pm$0.0060 & \underline{0.7524$\pm$0.0098} & 0.8804$\pm$0.0067 \\
&\fraudre & 0.5868$\pm$0.0208 & 0.7232$\pm$0.0182 & 0.6236$\pm$0.0178 & 0.7773$\pm$0.0140 & 0.6367$\pm$0.0316 & 0.8107$\pm$0.0197  \\
&\caregnn & 0.6151$\pm$0.0119 & 0.7290$\pm$0.0133 & 0.6859$\pm$0.0302 & 0.8223$\pm$0.0223 & 0.6943$\pm$0.0150 & 0.8316$\pm$0.0113 \\
&\pcgnn & 0.6335$\pm$0.0154 & 0.7412$\pm$0.0184 & 0.6950$\pm$0.0112 & 0.8239$\pm$0.0093 & 0.7202$\pm$0.0125 & 0.8495$\pm$0.0138 \\
&\bwgnnhomo & 0.5797$\pm$0.0183 & 0.7016$\pm$0.0213 & 0.6316$\pm$0.0280 & 0.7772$\pm$0.0173 & 0.6425$\pm$0.0604 & 0.8515$\pm$0.0103 \\
&\bwgnnhete & \underline{0.6558$\pm$0.0118} & \underline{0.7764$\pm$0.0196} & \underline{0.7137$\pm$0.0197} & \underline{0.8455$\pm$0.0146} & 0.7176$\pm$0.0705 & \underline{0.9026$\pm$0.0105} \\
&\ours(Ours) & \textbf{0.6884$\pm$0.0094} & \textbf{0.8279$\pm$0.0100} & \textbf{0.7462$\pm$0.0076} & \textbf{0.8833$\pm$0.0056} & \textbf{0.7988$\pm$0.0067} & \textbf{0.9233$\pm$0.0053} \\
\Xhline{\arrayrulewidth}
\multirow{\nummodels}{*}{\amazon} & \mlp & \textbf{0.9069$\pm$0.0084} & 0.9120$\pm$0.0241 & \textbf{0.9044$\pm$0.0083} & \underline{0.9524$\pm$0.0101} & 0.9114$\pm$0.0073 & 0.9695$\pm$0.0038 \\
&\gsage & 0.8999$\pm$0.0108 & 0.9095$\pm$0.0193 & 0.8966$\pm$0.0095 & \textbf{0.9549$\pm$0.0092} & \underline{0.9123$\pm$0.0065} & \textbf{0.9741$\pm$0.0031} \\
&\gat & 0.8685$\pm$0.0303 & 0.9126$\pm$0.0177 & 0.8874$\pm$0.0115 & 0.9475$\pm$0.0127 & 0.9023$\pm$0.0071 & 0.9640$\pm$0.0144 \\
&\gattwo & 0.8764$\pm$0.0218 & 0.9154$\pm$0.0125 & 0.8890$\pm$0.0127 & 0.9499$\pm$0.0100 & 0.9073$\pm$0.0086 & 0.9671$\pm$0.0059 \\
&\fraudre & 0.8188$\pm$0.0641 & 0.8906$\pm$0.0268 & 0.8423$\pm$0.0570 & 0.9193$\pm$0.0305 & 0.8596$\pm$0.0671 & 0.9442$\pm$0.0118  \\
&\caregnn & 0.9024$\pm$0.0058 & \textbf{0.9235$\pm$0.0245} & 0.8968$\pm$0.0074 & 0.9374$\pm$0.0085 & 0.9025$\pm$0.0071 & 0.9539$\pm$0.0085 \\
&\pcgnn & 0.8838$\pm$0.0297 & 0.9031$\pm$0.0206 & 0.8877$\pm$0.0094 & 0.9385$\pm$0.0112 & 0.8792$\pm$0.0137 & 0.9524$\pm$0.0065 \\
&\bwgnnhomo & 0.8650$\pm$0.0615 & 0.9102$\pm$0.0233 & 0.8857$\pm$0.0147 & 0.9419$\pm$0.0117 & 0.8891$\pm$0.0142 & 0.9700$\pm$0.0046 \\
&\bwgnnhete & 0.8024$\pm$0.0847 & 0.8759$\pm$0.0327 & 0.8680$\pm$0.0175 & 0.9458$\pm$0.0066 & 0.8791$\pm$0.0191 & 0.9692$\pm$0.0067 \\
&\ours(Ours) & \underline{0.9028$\pm$0.0133} & \underline{0.9172$\pm$0.0216} & \underline{0.9000$\pm$0.0060} & 0.9450$\pm$0.0068 & \textbf{0.9130$\pm$0.0082} & \underline{0.9701$\pm$0.0031}  \\
\Xhline{2\arrayrulewidth}
\end{tabular}
\end{table*}

\subsection{Relation-Attentive Aggregation}
\label{sec:relagg}
For each node, we have ($m$+1) representations at each layer as described above. \ours aggregates these representations using a dynamic attention with $N_{\beta}$ attention heads:
\begin{equation}
\vh_{i}^{(l+1)} = \sigma\left(\sum_{k=1}^{m+1} \beta_{ik}^{(l)} \mbP^{(l)} \vh_{i,k}^{(l)} \right)
\end{equation}
where $\mbP^{(l)}\in\mathbb{R}^{(d^{\prime}/N_{\beta})\times d^{\prime}}$ and $\beta_{ik}^{(l)}$ is computed by
\begin{equation}
\beta_{ik}^{(l)} = \dfrac{\text{exp}\left(\vba^{(l)} \sigma\left(\mbW^{(l)} [\vh_{i}^{(l)}||\text{ }\vh_{i,k}^{(l)}]\right)\right)}{\sum_{k'=1}^{m+1}\text{exp}\left(\vba^{(l)} \sigma\left(\mbW^{(l)} [\vh_{i}^{(l)}||\text{ }\vh_{i,k'}^{(l)}]\right)\right)}
\end{equation}
where $\mbW^{(l)}\in\mathbb{R}^{d^{\prime}\times 2d^{\prime}}$, and $\vba^{(l)}\in\mathbb{R}^{d^{\prime}}$ for $l=0,1,\cdots,L$. The attention coefficient $\beta_{ik}^{(l)}$ indicates the importance of the $k$-th relation for computing the representation of $v_i$ at the $l$-th layer. This attention coefficient can differ depending on nodes and layers.

\subsection{Aggregation with Multiple Layers}
\label{sec:hopagg}
It has been known that, when solving a node classification problem under heterophily~\cite{hetero}, it is helpful to explicitly consider the local and global neighbors by combining intermediate representations~\cite{jump}. To leverage this property, we aggregate representations from different layers:
\begin{equation}
\vh_{i}^{(L+1)} = \sigma\left(\sum_{l=0}^{L} \gammail \mtP \vh_{i}^{(l)} \right)
\end{equation}
with $N_{\gamma}$ heads, where $\mtP\in\mathbb{R}^{(d^{\prime}/N_{\gamma})\times d^{\prime}}$, $\vh_{i}^{(L+1)}\in\mathbb{R}^{d^\prime}$ is the final representation of $v_i$, and $\gammail$ is computed by
\begin{equation}
\gammail = \dfrac{\text{exp}\left(\vta \text{ } \sigma\left(\mtW [\vx_i\text{ }||\text{ }\vh_{i}^{(l)}]\right)\right)}{\sum_{l^\prime=0}^{L}\text{exp}\left(\vta \text{ } \sigma\left(\mtW [\vx_i\text{ }||\text{ }\vh_{i}^{(l^\prime)}]\right)\right)}
\end{equation}
where $\mtW\in\mathbb{R}^{d^{\prime}\times (d+d^{\prime})}$ and $\vta\in\mathbb{R}^{d^{\prime}}$. Note that the attention coefficient $\gammail$ is learned to imply the importance of the $l$-th layer's representation for computing the final representation of node $v_i$.

\subsection{Loss Function of \ours}
Using the final representation of each node, we predict the node label by computing $\widehat{y}_i=\mlp\left(\vh_{i}^{(L+1)}\right)$, where $\widehat{y}_i$ indicates $v_i$'s probability of being a fraud. The loss function of \ours is defined by $\sL_{\ours} = \sum_{v_i\in\sS} -y_i \text{log}\left(\widehat{y}_i\right)$ where $\sS\subset\sV$ is a training set.

\begin{table}[t]
\footnotesize
\caption{Dataset Statistic.}
\def\arraystretch{0.95}
\setlength{\tabcolsep}{0.47em}
\centering
\label{tb:data}
\begin{tabular}{cccccc}
\toprule
 & {\#}Nodes & {\#}Frauds & Relation & {\#}Edges & $d$ \\
\midrule
\multirow{4}{*}{\yelp} & \multirow{4}{*}{45,954} & \multirow{4}{*}{6,677 (14.53\%)} & R-U-R & 49,315 & \multirow{4}{*}{32} \\
& & & R-T-R & 573,616 & \\
& & & R-S-R & 3,402,743 & \\
& & & ALL & 3,846,979 & \\
\midrule
\multirow{4}{*}{\amazon} & \multirow{4}{*}{9,840} & \multirow{4}{*}{821 (8.34\%)} & U-P-U & 150,917 & \multirow{4}{*}{25} \\
& & & U-S-U & 2,979,223 & \\
& & & U-V-U & 838,682 & \\
& & & ALL & 3,627,491 & \\
\bottomrule
\vspace{-0.5cm}
\end{tabular}
\end{table}

\begin{figure*}[t]
	\centering
	\begin{subfigure}[h]{0.83\columnwidth}
		\centering
    	\includegraphics[width=\columnwidth]{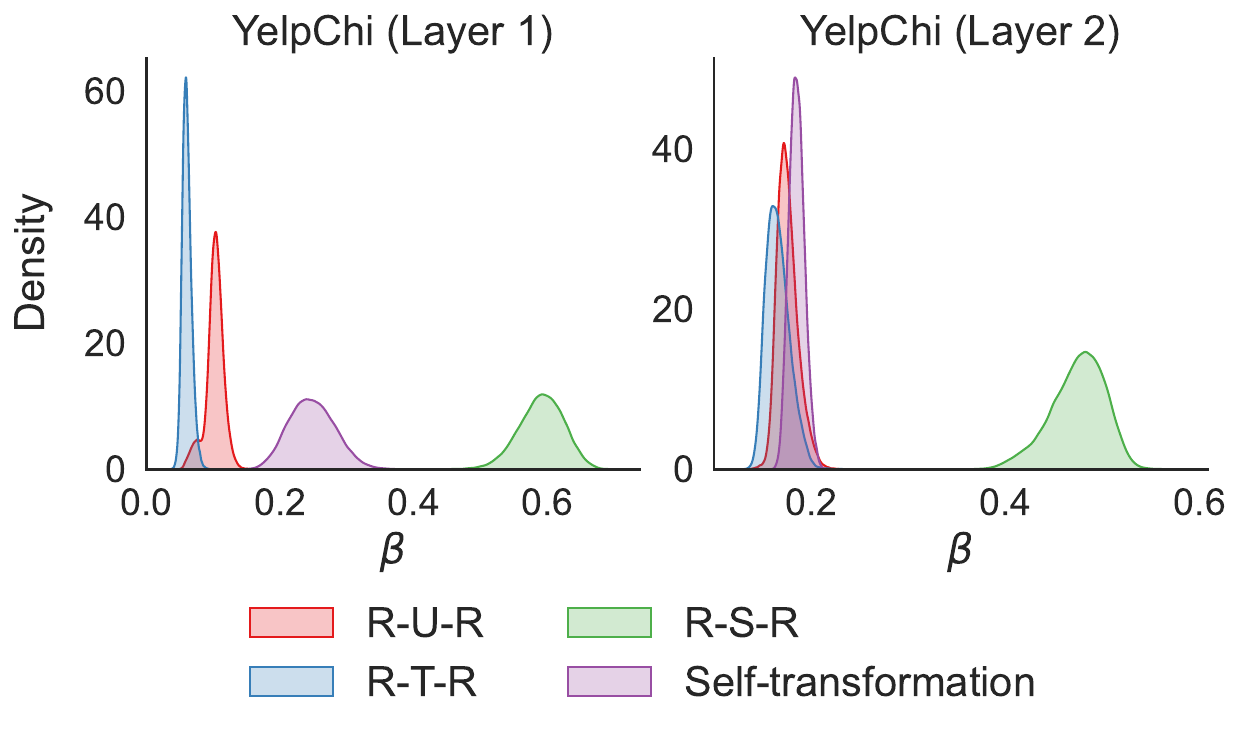}
    	\caption{\yelp}
    \end{subfigure}
	\begin{subfigure}[h]{1.17\columnwidth}
		\centering
    	\includegraphics[width=\columnwidth]{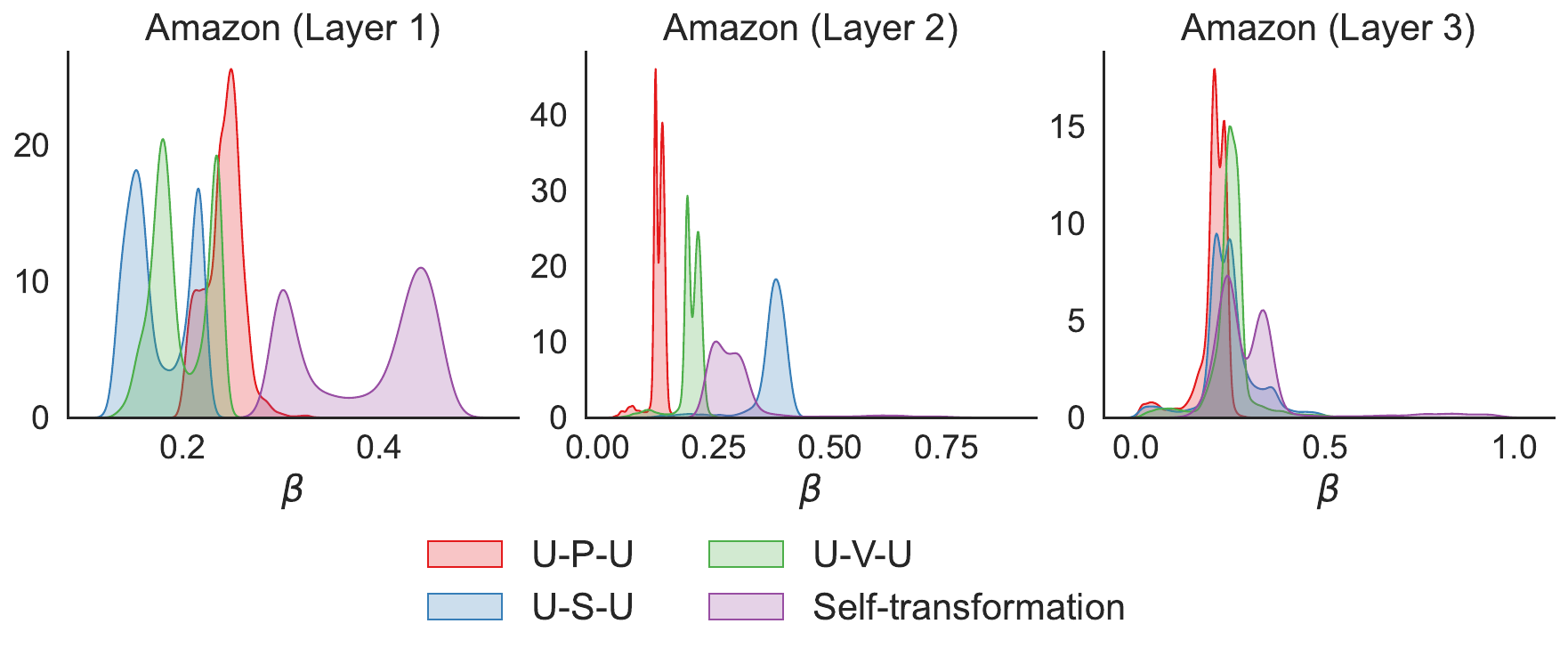}
    	\caption{\amazon}
    \end{subfigure}
    \caption{The kernel density estimate plots of $\beta_{ik}^{(l)}$ attention coefficients at each layer in \yelp ($L=2$) and \amazon ($L=3$).}
    \label{fig:relatt}
\end{figure*}

\section{Experiments}
We compare the performance of \ours with state-of-the-art fraud detection methods using real-world datasets. 

\subsection{Datasets and Experimental Setup}
To test a performance of a fraud detection method, we assume that we observe $p\%$ of the node labels and use them as a training set. Following a conventional setting~\cite{bwgnn}, we divide the remaining nodes into a validation set and a test set with a ratio of 1:2. We use two well-known real-world datasets, \yelp~\cite{yelp} and \amazon~\cite{amazon,amazonfeat}. In these datasets, there exist three different relations as shown in Table~\ref{tb:data}, where ALL indicates the total number of edges disregarding relations. Note that multiple edges between two nodes resulting from different relations are counted as one edge in ALL. In \amazon, we found that there are 2,104 duplicated nodes that do not have unique features. Since these nodes occur a data leaking issue, we removed these nodes.\footnote{In \amazon, the features include the feedback summary length, the entropy of ratings, the length of the username, and the ratio of helpful votes~\cite{amazonfeat}; the combination of these features should distinguish a node. It is very unlikely that multiple nodes share identical features, and thus we believe the duplicated nodes should be removed.}


We consider the following eight different baseline methods: \mlp, \gsage~\cite{gsage}, \gat~\cite{gat}, \gattwo~\cite{gat2}, \fraudre~\cite{fraudre}, \caregnn~\cite{caregnn}, \pcgnn~\cite{pcgnn}, and  \bwgnn~\cite{bwgnn}. We use the official implementations of these methods and the hyperparameters provided in the code or in the original paper. In \ours, we search the hyperparameters by considering the learning rate in $\{0.01, 0.001\}$, the weight decay in $\{0.001, 0.0001\}$, $L\in\{1,2,3\}$, the number of heads in $\{2,8\}$. We fix the batch size to be 1,024 and $d^\prime=64$ for all experiments. For all methods, we set the dimension of the final node representation to 64 and max epochs to 1,000 for fair comparisons. 

\subsection{Fraud Detection Performance}
In Table~\ref{tb:mainexp}, we show the fraud detection performance of the methods with different $p\in\{1,10,40\}$ in terms of two standard metrics, \fone and \auc. F1-macro is the unweighted mean of the F1 scores of two classes. AUC is the area under the ROC curve, representing the true positive rate against the false positive rate at various thresholds. We repeat all experiments ten times and report the average and the standard deviation. The best performance is boldfaced, and the second-best performance is underlined. In \yelp, we see that \ours significantly outperforms the baseline methods in all settings; \ours shows the best performance in terms of both \fone and \auc with three different ratios of labels. In \amazon, \ours shows comparable performance to the best-performing method.

\subsection{Qualitative Analysis \& Ablation Studies}
\begin{figure}[t]
	\centering
	\begin{subfigure}[h]{0.492\columnwidth}
		\centering
    	\includegraphics[width=\columnwidth]{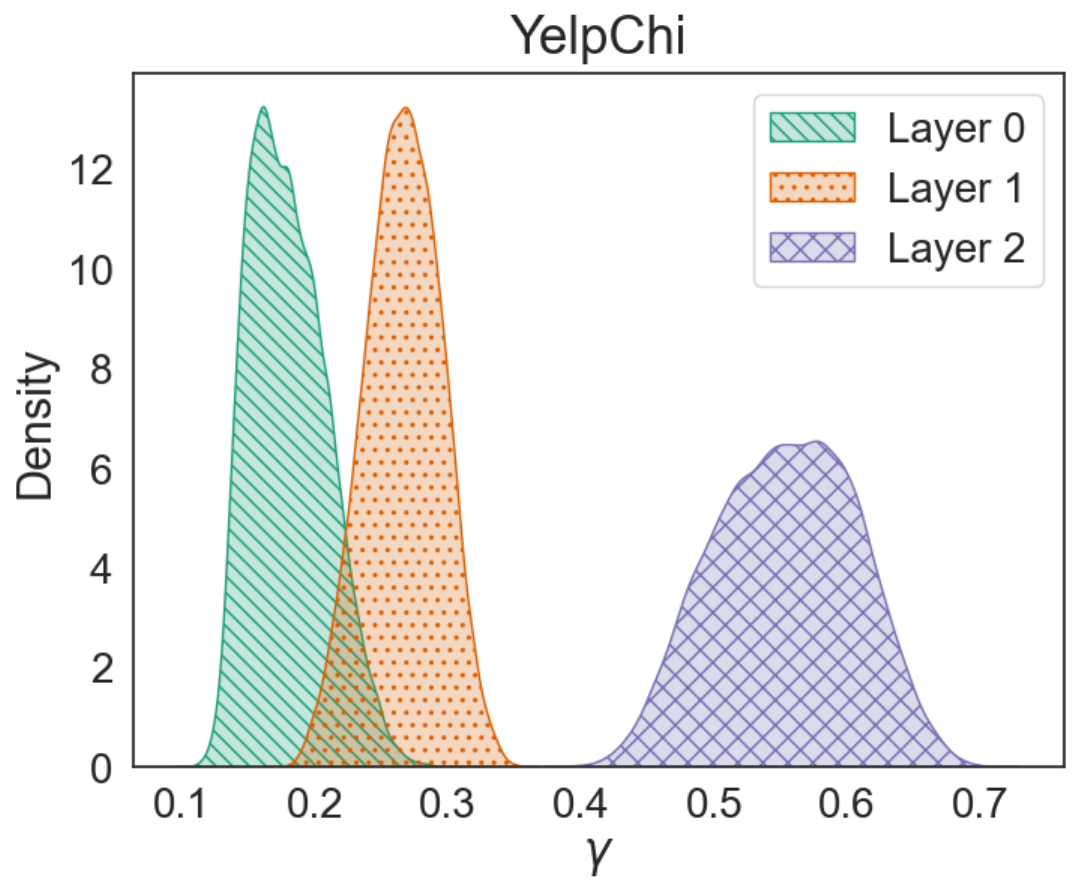}
    	\caption{\yelp}
    \end{subfigure}
	\begin{subfigure}[h]{0.4934\columnwidth}
		\centering
    	\includegraphics[width=\columnwidth]{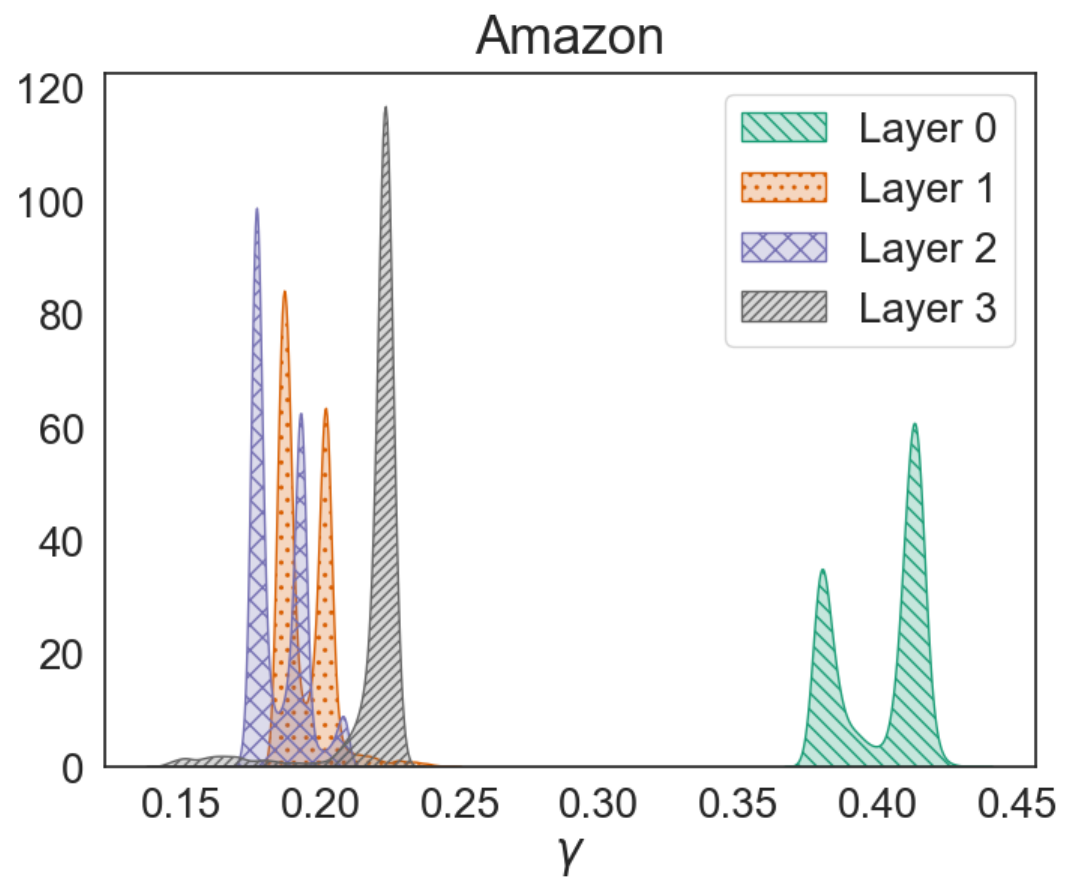}
    	\caption{\amazon}
    \end{subfigure}
    \caption{$\gammail$ attention coefficients in \yelp and \amazon.}
    \label{fig:layatt}
\end{figure}

As described in Section~\ref{sec:relagg}, \ours learns $\beta_{ik}^{(l)}$ which indicates the importance of $r_k$ to $v_i$ at the $l$-th layer. Figure~\ref{fig:relatt} shows the kernel density estimate plots of $\beta_{ik}^{(l)}$ values at each layer in \yelp with $L=2$ and \amazon with $L=3$. We see that each relation has a distinguishing distribution of the attention coefficients, which also differs between layers. Also, \ours learns $\gammail$ indicating the importance of the $l$-th layer representation for $v_i$ as described in Section~\ref{sec:hopagg}. Figure~\ref{fig:layatt} shows the distributions of $\gammail$. While Layer-2 tends to have a large importance in \yelp, Layer-0 has a relatively large importance in \amazon. In Figure~\ref{fig:relatt} and Figure~\ref{fig:layatt}, we see that the attention coefficient values are not concentrated on specific values, and some of their distributions are multimodal. This shows that our dynamic graph attention mechanism works as expected, resulting in various attention coefficients depending on target nodes.

We conduct ablation studies by disregarding relation types (w/o rel. types),  by removing the layer aggregation (w/o layer agg.), and by utilizing only a single layer instead of multiple layers (w/ single layer) in \ours as presented in Table~\ref{tb:ablation}. We see that the performance of \ours significantly degrades by dropping either the relation-attentive aggregation or the layer-attentive aggregation, which indicates that both of these play a critical role in \ours. In addition, we observe that considering higher-order neighbors via multiple layers helps increase fraud detection performance, mainly when enough labels are provided.



\begin{table}[t]
\renewcommand{\arraystretch}{1.1}
\footnotesize
\caption{Ablation studies of \ours. \auc scores on \yelp using different percentages of labels are reported.}
\setlength{\tabcolsep}{0.84em}
\centering
\label{tb:ablation}
\begin{tabular}{c|ccc}
\Xhline{\arrayrulewidth}
& {1\%} & {10\%} & {40\%} \\
\Xhline{\arrayrulewidth}
\ours & 0.8279$\pm$0.0100 & 0.8833$\pm$0.0056 & 0.9233$\pm$0.0053 \\
w/o rel. types & 0.7200±0.0134 & 0.8079$\pm$0.0134 & 0.8716$\pm$0.0054 \\
w/o layer agg. &  0.7153$\pm$0.1349 & 0.8377$\pm$0.1128 & 0.8775$\pm$0.1260 \\
w/ single layer & 0.8214$\pm$0.0085 & 0.8790$\pm$0.0085 & 0.9076$\pm$0.0087 \\
\Xhline{\arrayrulewidth}
\end{tabular}
\vspace{-0.3cm}
\end{table}

\section{Conclusion \& Future Work}
We propose a dynamic attention-based fraud detection, performing relation-wise and layer-wise attentive aggregations. The learnable attention coefficients allow \ours to concentrate more on neighbors, relations, and layers beneficial to predict the label of the target node. By dynamically adapting the attention coefficients for individual nodes, this attention mechanism is especially effective in fraud detection on graphs with heterophily.

In future work, we will investigate our attention mechanisms from a theoretical point of view. Specifically, we will explore how the attention coefficients are learned under a specific heterophily property. Moreover, we plan to extend our approaches and methods to more complex relational graphs~\cite{hynt, bive}. Also, we will extend \ours to handle evolving graphs where new nodes appear and new edges are formed over time~\cite{ingram}.


\bibliographystyle{IEEEtran}
\bibliography{IEEEabrv,ref_drag}

\end{document}